\documentclass[10pt,twocolumn,letterpaper]{article}

\usepackage{cvpr}
\usepackage{times}
\usepackage{epsfig}
\usepackage{graphicx}
\usepackage{amsmath}
\usepackage{amssymb}
\usepackage{tabulary}
\usepackage{multirow}

\newcolumntype{x}[1]{>{\centering\arraybackslash}p{#1pt}}

\newlength\savewidth\newcommand\shline{\noalign{\global\savewidth\arrayrulewidth
  \global\arrayrulewidth 1pt}\hline\noalign{\global\arrayrulewidth\savewidth}}
\newcommand{\tablestyle}[2]{\setlength{\tabcolsep}{#1}\renewcommand{\arraystretch}{#2}\centering\footnotesize}

\usepackage[pagebackref=true,breaklinks=true,letterpaper=true,colorlinks,bookmarks=false]{hyperref}

\cvprfinalcopy 

\def\cvprPaperID{****} 

\begin{document}

\title{Deep Snake for Real-Time Instance Segmentation}

\author{
Sida Peng$^1$
\quad
Wen Jiang$^1$
\quad
Huaijin Pi$^1$
\quad
Xiuli Li$^2$
\quad
Hujun Bao$^{1}$
\quad
Xiaowei Zhou$^1$\thanks{The authors from Zhejiang University are affiliated with the State Key Lab of CAD\&CG. Corresponding author: Xiaowei Zhou.} \\[2mm]
$^1$Zhejiang University
\quad
$^2$Deepwise AI Lab
}

\maketitle

\begin{abstract}
    This paper introduces a novel contour-based approach named deep snake for real-time instance segmentation. Unlike some recent methods that directly regress the coordinates of the object boundary points from an image, deep snake uses a neural network to iteratively deform an initial contour to match the object boundary, which implements the classic idea of snake algorithms with a learning-based approach. For structured feature learning on the contour, we propose to use circular convolution in deep snake, which better exploits the cycle-graph structure of a contour compared against generic graph convolution. Based on deep snake, we develop a two-stage pipeline for instance segmentation: initial contour proposal and contour deformation, which can handle errors in object localization. Experiments show that the proposed approach achieves competitive performances on the Cityscapes, KINS, SBD and COCO datasets while being efficient for real-time applications with a speed of 32.3 fps for 512$\times$512 images on a 1080Ti GPU. The code is available at \href{https://github.com/zju3dv/snake/}{https://github.com/zju3dv/snake/}.


\end{abstract}

\section{Introduction}

Instance segmentation is the cornerstone of many computer vision tasks, such as video analysis, autonomous driving, and robotic grasping, which require both accuracy and efficiency. Most of the state-of-the-art instance segmentation methods \cite{he2017mask, liu2018path, chen2019hybrid, huang2019mask} perform pixel-wise segmentation within a bounding box given by an object detector \cite{ren2015faster}, which may be sensitive to the inaccurate bounding box. Moreover, representing an object shape as dense binary pixels generally results in costly post-processing.

An alternative shape representation is the object contour, which is a set of vertices along the object silhouette. In contrast to pixel-based representation, a contour is not limited within a bounding box and has fewer parameters. Such a contour-based representation has long been used in image segmentation since the seminal work by Kass et al. \cite{kass1988snakes}, which is well known as snakes or active contours. Given an initial contour, the snake algorithm iteratively deforms it to match the object boundary by optimizing an energy functional defined with low-level features, such as image intensity or gradient. While many variants \cite{cohen1991active, cootes1995active, gunn1997robust} have been developed in literature, these methods are prone to local optima as the objective functions are handcrafted and typically nonconvex.

\begin{figure}[t]
\centering
\includegraphics[width=1\linewidth]{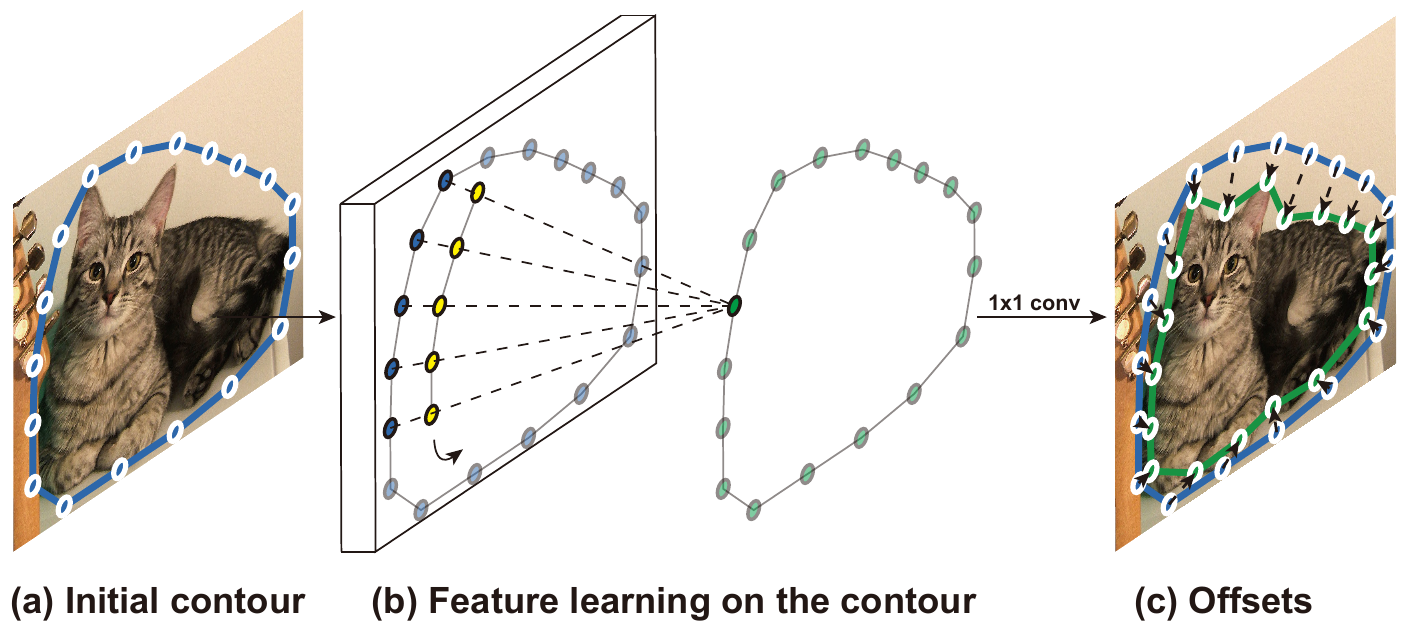}
\caption{\textbf{The basic idea of deep snake.} Given an initial contour, image features are extracted at each vertex (a). Since the contour is a cycle graph, circular convolution is applied for feature learning on the contour (b). The blue, yellow and green nodes denote the input features, the kernel of circular convolution, and the output features, respectively. Finally, offsets are regressed at each vertex to deform the contour to match the object boundary (c).}
\label{fig:basic_idea}
\vspace{-4mm}
\end{figure}

Some recent learning-based segmentation methods \cite{jetley2017straight, xu2019explicit, xie2020polarmask} also represent objects as contours and try to directly regress the coordinates of contour vertices from an RGB image. Although such methods are fast, most of them do not perform as well as pixel-based methods. Instead, Ling et al. \cite{ling2019fast} adopt the deformation pipeline of traditional snake algorithms and train a neural network to evolve an initial contour to match the object boundary. Given a contour with image features, it regards the input contour as a graph and uses a graph convolutional network (GCN) to predict vertex-wise offsets between contour points and the target boundary points. It achieves competitive accuracy compared with pixel-based methods while being much faster. However, the method proposed in \cite{ling2019fast} is designed to help annotation and lacks a complete pipeline for automatic instance segmentation. Moreover, treating the contour as a general graph with a generic GCN does not fully exploit the special topology of a contour.

In this paper, we propose a learning-based snake algorithm, named deep snake, for real-time instance segmentation. Inspired by previous methods \cite{kass1988snakes, ling2019fast}, deep snake takes an initial contour as input and deforms it by regressing vertex-wise offsets. Our innovation is introducing the circular convolution for efficient feature learning on a contour, as illustrated in Figure~\ref{fig:basic_idea}. We observe that the contour is a cycle graph that consists of a sequence of vertices connected in a closed cycle. Since every vertex has the same degree equal to two, we can apply the standard 1D convolution on the vertex features. Considering that the contour is periodic, deep snake introduces the circular convolution, which indicates that an aperiodic function (1D kernel) is convolved in the standard way with a periodic function (features defined on the contour). The kernel of circular convolution encodes not only the feature of each vertex but also the relationship among neighboring vertices. In contrast, the generic GCN performs pooling to aggregate information from neighboring vertices. The kernel function in our circular convolution amounts to a learnable aggregation function, which is more expressive and results in better performance than using a generic GCN, as demonstrated by our experimental results in Section~\ref{section:ablation}.

Based on deep snake, we develop a pipeline for instance segmentation. Given an initial contour, deep snake can iteratively deform it to match the object boundary and obtain the object shape. The remaining question is how to initialize a contour, whose importance has been demonstrated in classic snake algorithms. Inspired by \cite{papadopoulos2017extreme, maninis2018deep, zhou2019bottom}, we propose to generate an octagon formed by object extreme points as the initial contour, which generally encloses the object tightly. Specifically, we integrate deep snake with an object detector. The detected bounding box  initializes a diamond contour defined by four center points on the edges. Then, deep snake takes the diamond as input and outputs offsets that point from diamond vertices to object extreme points, which are used to construct an octagon following \cite{zhou2019bottom}. Finally, deep snake deforms the octagon contour to match the object boundary.

Our approach exhibits competitive performances on Cityscapes \cite{cordts2016cityscapes}, KINS \cite{qi2019amodal}, SBD \cite{hariharan2011semantic} and COCO \cite{lin2014microsoft} datasets, while being efficient for real-time instance segmentation, 32.3 fps for $512 \times 512$ images on a GTX 1080ti GPU. The following two facts make learning-based snake fast and accurate. First, our approach can deal with errors in the object localization stage and thus allows a light detector. Second, the contour representation has fewer parameters than the pixel-based representation and does not require costly post-processing, e.g., mask upsampling.

In summary, this work has the following contributions:

\begin{itemize}
\item We propose a learning-based snake algorithm for real-time instance segmentation and introduce the circular convolution for feature learning on the contour.
\item We propose a two-stage pipeline for instance segmentation: initial contour proposal and contour deformation. Both stages can deal with errors in the initial object localization.
\item We demonstrate state-of-the-art performances of our approach on Cityscapes, KINS, SBD and COCO datasets. For $512 \times 512$ images, our algorithm runs at 32.3 fps, which is efficient for real-time applications.
\end{itemize}

\section{Related work}

\paragraph{Pixel-based methods.} Most methods \cite{dai2016instance, li2017fully, he2017mask, liu2018path} perform instance segmentation on the pixel level within a region proposal, which works particularly well with standard CNNs. A representative instantiation is Mask R-CNN \cite{he2017mask}. It first detects objects and then uses a mask predictor to segment instances within the proposed boxes. To better exploit the spatial information inside the box, PANet \cite{liu2018path} fuses mask predictions from fully-connected layers and convolutional layers. Such proposal-based approaches achieve state-of-the-art performance. One limitation of these methods is that they cannot resolve errors in localization, such as too small or shifted boxes. In contrast, our approach deforms the detected boxes to the object boundaries, so the spatial extension of object shapes will not be limited.

There exist some pixel-based methods \cite{bai2017deep, neven2019instance, liu2018affinity, gao2019ssap, yang2019dense} that are free of region proposals. In these methods, every pixel produces the auxiliary information, and then a clustering algorithm groups pixels into object instances based on their information. The auxiliary information and grouping algorithms could be various. \cite{bai2017deep} predicts the boundary-aware energy for each pixel and uses the watershed transform algorithm for grouping. \cite{neven2019instance} differentiates instances by learning instance-level embeddings. \cite{liu2018affinity, gao2019ssap} consider the input image as a graph and regress pixel affinities, which are then processed by a graph merge algorithm. Since the mask is composed of dense pixels, the post-clustering algorithms tend to be time-consuming.

\paragraph{Contour-based methods.} In these methods, the object shape comprises a sequence of vertices along the object boundary. Traditional snake algorithms \cite{kass1988snakes, cohen1991active, cootes1995active, gunn1997robust} first introduced the contour-based representation for image segmentation. They deform an initial contour to the object boundary by optimizing a handcrafted energy with respect to the contour coordinates. To improve the robustness of these methods, \cite{marcos2018learning} proposed to learn the energy function in a data-driven manner. Instead of iteratively optimizing the contour, some recent learning-based methods \cite{jetley2017straight, xu2019explicit} try to regress the coordinates of contour points from an RGB image, which is much faster. However, their reported accuracy is not on par with state-of-the-art pixel-based methods.

In the field of semi-automatic annotation, \cite{castrejon2017annotating, acuna2018efficient, ling2019fast} have tried to perform the contour labeling using other networks instead of standard CNNs. \cite{castrejon2017annotating, acuna2018efficient} predict the contour points sequentially using a recurrent neural network. To avoid sequential inference, \cite{ling2019fast} follows the pipeline of snake algorithms and uses a graph convolutional network to predict vertex-wise offsets for contour deformation. This strategy significantly improves the annotation speed while being as accurate as pixel-based methods. However, \cite{ling2019fast} lacks a pipeline for instance segmentation and does not fully exploit the special topology of a contour. Instead of treating the contour as a general graph, deep snake leverages the cycle graph topology and introduces the circular convolution for efficient feature learning on a contour.

\section{Proposed approach}

Inspired by \cite{kass1988snakes, ling2019fast}, we perform object segmentation by deforming an initial contour to match  object boundary. Specifically, deep snake takes a contour as input and predicts per-vertex offsets pointing to the object boundary. Features on contour vertices are extracted from the input image with a CNN backbone. To fully exploit the contour topology, we propose the circular convolution for efficient feature learning on the contour, which facilitates deep snake to learn the deformation. Based on deep snake, we also develop a pipeline for instance segmentation.

\subsection{Learning-based snake algorithm}

Given an initial contour, traditional snake algorithms treat the coordinates of the vertices as a set of variables and optimize an energy functional with respect to these variables. By designing proper forces at the contour coordinates, the algorithms could drive the contour to the object boundary. However, since the energy functional is typically nonconvex and handcrafted based on low-level image features, the optimization tends to find local optimal solutions.

In contrast, deep snake directly learns to evolve the contour in an end-to-end manner. Given a contour with $N$ vertices $\{\mathbf{x}_i | i = 1, ..., N\}$, we first construct feature vectors for each vertex. The input feature $f_i$ for a vertex $\mathbf{x}_i$ is a concatenation of learning-based features and the vertex coordinate: $[F(\mathbf{x}_i); \mathbf{x}_i]$, where $F$ denotes the feature maps. 
The feature maps $F$ are obtained by applying a CNN backbone on the input image. The CNN backbone is shared with the detector in our instance segmentation pipeline, which will be discussed later. The image feature $F(\mathbf{x}_i)$ is computed using the bilinear interpolation at the vertex coordinate $\mathbf{x}_i$. The appended vertex coordinate is used to encode the spatial relationship among contour vertices. Since the deformation should not be affected by the translation of the contour in the image, we subtract each dimension of $\mathbf{x}_i$ by the minimum value over all vertices.

\begin{figure}[t]
\begin{minipage}[c]{0.45\linewidth}
\includegraphics[width=\textwidth]{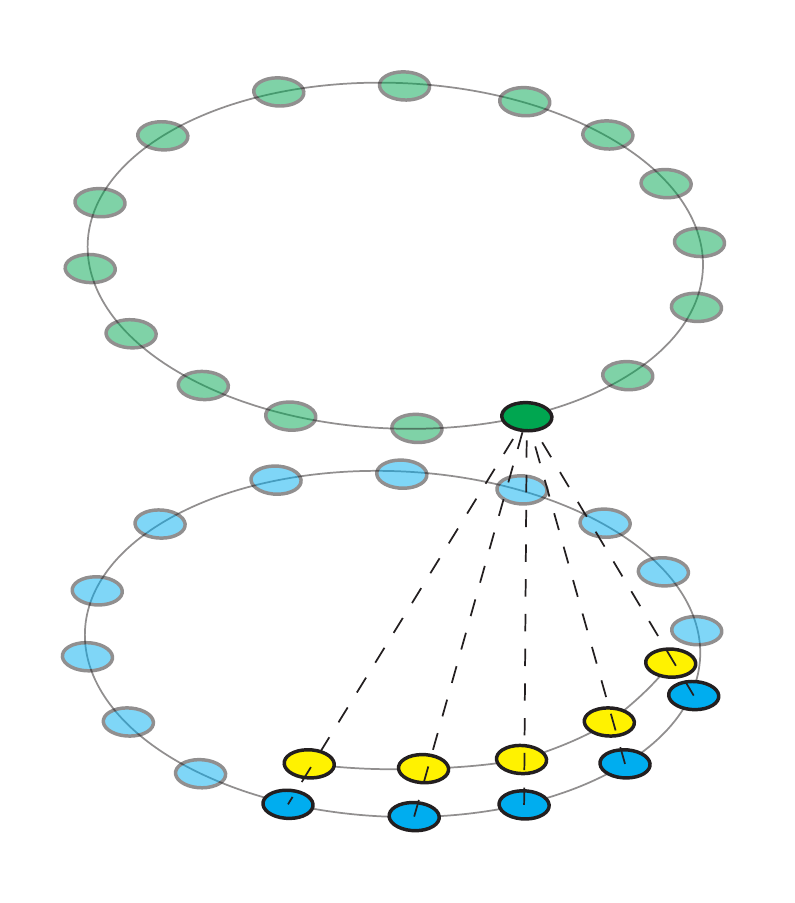}
\end{minipage}\hfill
\begin{minipage}[c]{0.55\linewidth}
    \caption{\footnotesize \textbf{Circular Convolution.} The blue nodes are the input features defined on a contour, the yellow nodes represent the kernel function, and the green nodes are the output features. The highlighted green node is the inner product between the kernel function and the highlighted blue nodes, which is the same as the standard convolution. The output features of circular convolution have the same length as the input features.}
\label{fig:circular_convolution}
\end{minipage}\vspace{-5mm}
\end{figure}

Given the input features defined on a contour, deep snake introduces the circular convolution for the feature learning, as illustrated in Figure~\ref{fig:circular_convolution}. In general, the features of contour vertices can be treated as a 1-D discrete signal $f: \mathbb{Z} \to \mathbb{R}^D$ and processed by the standard convolution. But this breaks the topology of the contour. Therefore, we extend $f$ to be a periodic signal defined as:
\begin{equation}
    (f_N)_i \triangleq \sum_{j = -\infty}^{\infty} f_{i - jN}, 
\end{equation}
and propose to encode the periodic features by the circular convolution defined as: 
\begin{equation}
    (f_N \ast k)_i = \sum_{j = -r}^r (f_N)_{i + j}k_j,
\end{equation}
where $k: [-r, r] \to \mathbb{R}^D$ is a learnable kernel function and the operator $\ast$ is the standard convolution. 

Similar to the standard convolution, we can construct a network layer based on the circular convolution for feature learning, which is easy to be integrated into a modern network architecture. After the feature learning, deep snake applies three 1$\times$1 convolution layers to the output features for each vertex and predicts vertex-wise offsets between contour points and the target points, which are used to deform the contour. In all experiments, the kernel size of circular convolution is fixed to be nine.

As discussed in the introduction, the proposed circular convolution better exploits the circular structure of the contour than the generic graph convolution. We will show the experimental comparison in Section \ref {section:ablation}. An alternative method is to use standard CNNs to regress a pixel-wise vector field from the input image to guide the evolution of the initial contour \cite{rupprecht2016deep, peng2019pvnet, wang2019object}. We argue that an important advantage of deep snake over the standard CNNs is the object-level structured prediction, i.e., the offset prediction at a vertex depends on other vertices of the same contour. Therefore, deep snake will predict a more reasonable offset for a vertex located far from the object. Standard CNNs may have difficulty in this case, as the regressed vector field may drive this vertex to another object which is closer. 


\begin{figure*}[t]
\centering
\includegraphics[width=1\linewidth]{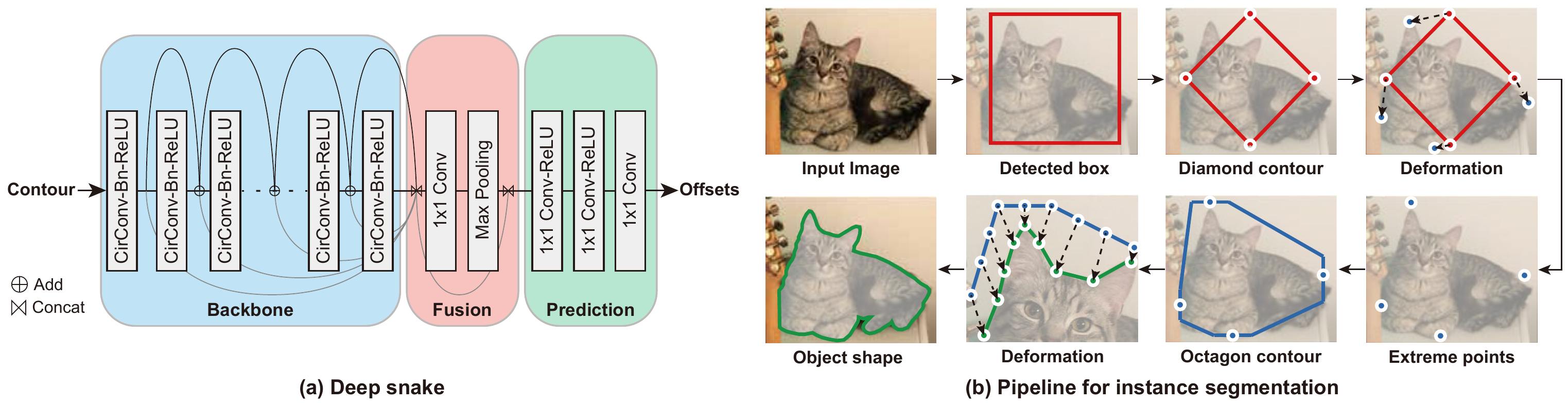}
\caption{\textbf{Proposed contour-based model for instance segmentation.} (a) Deep snake consists of three parts: a backbone, a fusion block, and a prediction head. It takes a contour as input and outputs vertex-wise offsets to deform the contour. (b) Based on deep snake, we propose a two-stage pipeline for instance segmentation: initial contour proposal and contour deformation. The box proposed by the detector gives a diamond contour, whose four vertices are then shifted to object extreme points by deep snake. An octagon is constructed based on the extreme points. Taking the octagon as the initial contour, deep snake iteratively deforms it to match the object boundary.}
\label{fig:architecture}
\vspace{-2mm}
\end{figure*}

\paragraph{Network architecture.}

\label{section:snakenet}

Figure~\ref{fig:architecture}(a) shows the detailed schematic. Following ideas from \cite{qi2017pointnet, wang2018dynamic, li2019can}, deep snake consists of three parts: a backbone, a fusion block, and a prediction head. The backbone is comprised of 8 ``CirConv-Bn-ReLU'' layers and uses residual skip connections for all layers, where ``CirConv'' means circular convolution. The fusion block aims to fuse the information across all contour points at multiple scales. It concatenates features from all layers in the backbone and forwards them through a 1$\times$1 convolution layer followed by max pooling. The fused feature is then concatenated with the feature of each vertex. The prediction head applies three 1$\times$1 convolution layers to the vertex features and output vertex-wise offsets.

\subsection{Deep snake for instance segmentation}

\label{section:instance_segmentation}

Figure~\ref{fig:architecture}(b) overviews the proposed pipeline for instance segmentation. We combine deep snake with an object detector. The detector first produces object bounding boxes that are used to construct diamond contours. Then deep snake shifts the diamond vertices to object extreme points, which are used to construct octagon contours. Finally, deep snake takes octagons as initial contours and performs iterative contour deformation to obtain the object shape.

\paragraph{Initial contour proposal.}

\label{section:initial_proposal}

Most active contour models require precise initial contours. Since the octagon proposed in \cite{zhou2019bottom} tightly encloses the object, we choose it as the initial contour, as shown in Figure~\ref{fig:architecture}(b). This octagon is formed by four extreme points, which are top, leftmost, bottom, rightmost pixels in an object, respectively, denoted by $\{\mathbf{x}^{ex}_i | i = 1, 2, 3, 4\}$. Given a detected object box, we extract four center points at the top, left, bottom, right box edges, denoted by $\{\mathbf{x}^{bb}_i | i = 1, 2, 3, 4\}$, and then connect them to get a diamond contour. Deep snake takes this contour as input and outputs four offsets that point from each vertex $\mathbf{x}^{bb}_i$ to the extreme point $\mathbf{x}^{ex}_i$, namely $\mathbf{x}^{ex}_i - \mathbf{x}^{bb}_i$. In practice, to consider more context information, the diamond contour is uniformly upsampled to 40 points, and deep snake correspondingly outputs 40 offsets. The loss function only supervises the offsets at $\mathbf{x}^{bb}_i$.

We construct the octagon by generating four line segments based on extreme points and connecting their endpoints. Specifically, the four extreme points define a new bounding box. From each extreme point, a line is extended along the corresponding box edge in both directions by 1/4 of the edge length and truncated if it meets the box corner. Then, the endpoints of the four line segments are connected to form the octagon.

\paragraph{Contour deformation.} We first uniformly sample $N$ points along the octagon contour starting from the top extreme points $\mathbf{x}^{ex}_1$. Similarly, the ground-truth contour is generated by uniformly sampling $N$ vertices along the object boundary and defining the first vertex as the one nearest to $\mathbf{x}^{ex}_1$. Deep snake takes the initial contour as input and outputs $N$ offsets that point from each vertex to the target boundary point. We set $N$ as $128$ in all experiments, which can uniformly cover most object shapes.

However, regressing the offsets in one pass is challenging, especially for vertices far away from the object. Inspired by \cite{kass1988snakes, ling2019fast, wang2018pixel2mesh}, we deal with this problem in an iterative optimization fashion. Specifically, our approach first predicts $N$ offsets based on the current contour and then deforms this contour by vertex-wise adding the offsets to its vertex coordinates. The deformed contour can be used for the next iteration. In experiments, the number of inference iteration is set as 3 unless otherwise stated.

Note that the contour is an alternative representation for the spatial extension of an object. By deforming the initial contour to match the object boundary, deep snake could address the localization errors from the detector.

\begin{figure}[t]
\centering
\includegraphics[width=1\linewidth]{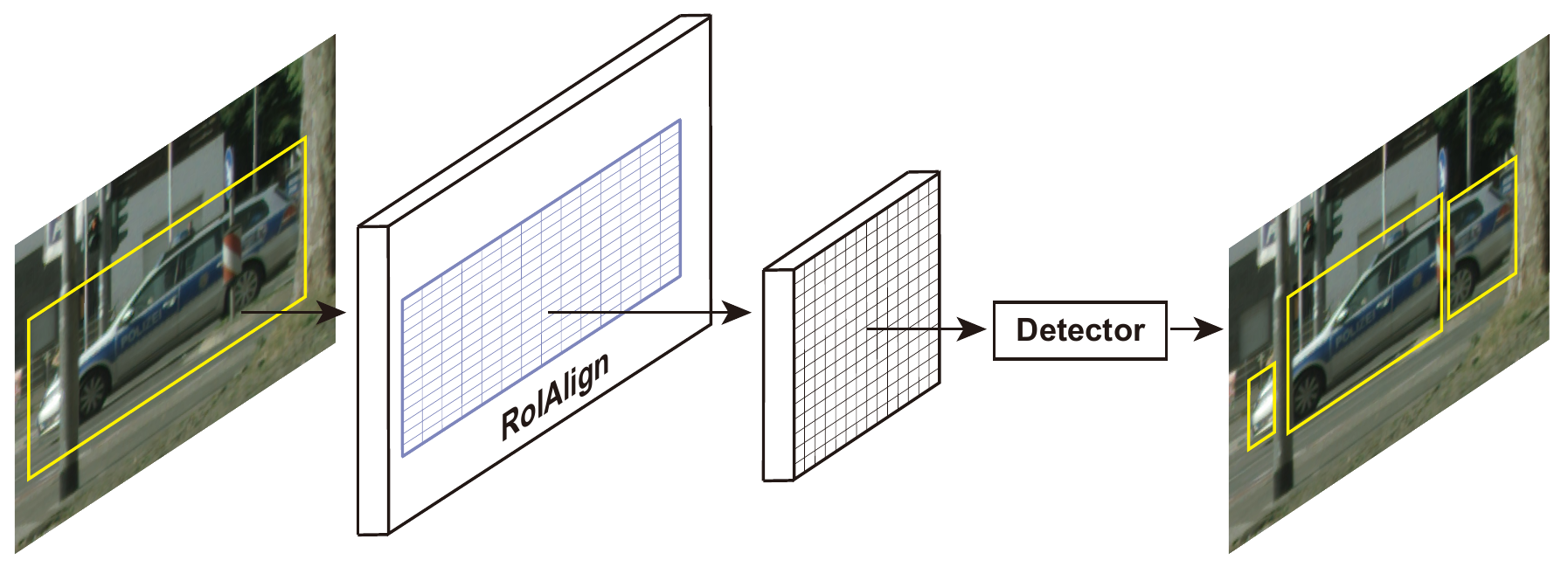}
\caption{\textbf{Multi-component detection.} Given an object box, we perform RoIAlign to obtain the feature map and use a detector to detect the component boxes.}
\label{fig:component}
\vspace{-2mm}
\end{figure}

\paragraph{Multi-component detection.} Some objects are split into several components due to occlusions, as shown in Figure~\ref{fig:component}. However, a contour can only outline one component. To overcome this problem, we propose to use another detector to find the object components within the object box. Figure~\ref{fig:component} shows the basic idea. Specifically, using the detected box, our approach performs RoIAlign \cite{he2017mask} to extract a feature map and adds a detector branch on the feature map to produce the component boxes. 
For the detected components, we use deep snake to segment each of them and then merge the segmentation results.

\section{Implementation details}

\paragraph{Training strategy.} For the training of deep snake, we use the smooth $\ell_1$ loss proposed in \cite{girshick2015fast} to learn the two deformation processes. The loss function for extreme point prediction is defined as
\begin{equation}
    L_{ex} = \frac{1}{4} \sum\limits_{i=1}^4 \ell_1(\tilde{\mathbf{x}}^{ex}_i - \mathbf{x}^{ex}_i),
\end{equation}
where $\tilde{\mathbf{x}}^{ex}_i$ is the predicted extreme point. And the loss function for iterative contour deformation is defined as
\begin{equation}
    L_{iter} = \frac{1}{N} \sum\limits_{i=1}^N \ell_1(\tilde{\mathbf{x}}_i - \mathbf{x}^{gt}_i),
\end{equation}
where $\tilde{\mathbf{x}}_i$ is the deformed contour point and $\mathbf{x}^{gt}_i$ is the ground-truth boundary point. For the detection part, we adopt the same loss function as the original detection model. The training details change with datasets, which will be described in Section~\ref{section:performance}.

\paragraph{Detector.} We adopt CenterNet \cite{zhou2019objects} as the detector for all experiments. 
CenterNet reformulates the detection task as a keypoint detection problem and achieves an impressive trade-off between speed and accuracy. For the object box detector, we adopt the same setting as \cite{zhou2019objects}, which outputs class-specific boxes. For the component box detector, a class-agnostic CenterNet is adopted. Specifically, given an $H \times W \times C$ feature map, the class-agnostic CenterNet outputs an $H \times W \times 1$ tensor representing the component center and an $H \times W \times 2$ tensor representing the box size.

\section{Experiments}


\subsection{Datasets and Metrics}

\paragraph{Cityscapes \cite{cordts2016cityscapes}}
contains $2,975$ training, 500 validation and $1,525$ testing images with high quality annotations. Besides, it has 20k images with coarse annotations.
The performance is evaluated in terms of the average precision (AP) metric averaged over eight semantic classes of the dataset.

\paragraph{KINS \cite{qi2019amodal}} was created by additionally annotating Kitti \cite{geiger2013vision} dataset with instance-level semantic annotation.
This dataset is used for amodal instance segmentation, which aims to recover complete instance shapes even under occlusion.
KINS consists of $7,474$ training images and $7,517$ testing images.
Following its setting, we evaluate our approach on seven object categories in terms of the AP metric.

\paragraph{SBD \cite{hariharan2011semantic}}
re-annotates $11,355$ images from the PASCAL VOC \cite{everingham2010pascal} dataset with instance-level boundaries. The reason that we don't evaluate on PASCAL VOC is that its annotations contain holes, which is not suitable for contour-based methods. SBD is split into $5,623$ training images and $5,732$ testing images. We report our results in terms of 2010 VOC AP$_{vol}$ \cite{hariharan2014simultaneous}, AP$_{50}$, AP$_{70}$ metrics. AP$_{vol}$ is the average of AP with nine thresholds from 0.1 to 0.9.

\paragraph{COCO \cite{lin2014microsoft}} is one of the most challenging datasets for instance segmentation. It consists of 115k training , 5k validation and 20k testing images. We report our results in terms of the AP metric.

\subsection{Ablation studies}

\label{section:ablation}

We conduct ablation studies on the SBD dataset as it has 20 semantic categories which could fully evaluate the ability to handle various object shapes. The three proposed components are evaluated, including our network architecture, initial contour proposal, and circular convolution. In these experiments, the detector and deep snake are trained end-to-end for 160 epochs with multi-scale data augmentation. The learning rate starts from $1e^{-4}$ and decays by half at 80 and 120 epochs. 

Table~\ref{table:ablation} summarizes the results of ablation studies. The row ``Baseline'' lists the result of a direct combination of Curve-gcn \cite{ling2019fast} with CenterNet \cite{zhou2019objects}. Specifically, the detector produces object boxes, which gives ellipses around objects. Then ellipses are deformed towards object boundaries through Graph-ResNet. Note that, this baseline method represents the contour as a graph and uses a graph convolution network for contour deformation.

\begin{table}
\tablestyle{4pt}{1.05}
\begin{tabular}{l|x{42}|x{42}|x{42}}
& AP$_{vol}$ & AP$_{50}$ & AP$_{70}$ \\[.1em]
\shline
Baseline & 50.9 & 58.8 & 43.5 \\
+ Architecture & 52.3 & 59.7 & 46.0 \\
+ Initial proposal & 53.6 & 61.1 & 47.6 \\
+ Circular convolution & \textbf{54.4} & \textbf{62.1} & \textbf{48.3} \\
\end{tabular}\vspace{1mm}
\caption{\textbf{Ablation studies on SBD val set .} The baseline is a direct combination of Curve-gcn \cite{ling2019fast} and CenterNet \cite{zhou2019objects}. The second model reserves the graph convolution and replaces the network architecture with our proposed one, which yields 1.4 AP$_{vol}$ improvement. Then we add the initial contour proposal before contour deformation, which improves AP$_{vol}$ by 1.3. The fourth row shows that replacing graph convolution with circular convolution further yields 0.8 AP$_{vol}$ improvement.}
\vspace{-3mm}
\label{table:ablation}
\end{table}

\begin{table}
\tablestyle{4pt}{1.05}
\begin{tabular}{l|x{28}|x{28}|x{28}|x{28}|x{28}}
    & Iter. 1 & Iter. 2 & Iter. 3 & Iter. 4 & Iter. 5 \\[.1em]
\shline
Graph conv & 50.2 & 51.5 & 53.6 & 52.2 & 51.6 \\
Circular conv & 50.6 & 54.2 & \textbf{54.4} & 54.0 & 53.2 \\
\end{tabular}\vspace{1mm}
\caption{\textbf{Results of models with different convolution operators and different iterations} on SBD in terms of the AP$_{vol}$ metric. Circular convolution outperforms graph convolution across all inference iterations. Furthermore, circular convolution with two iterations outperforms graph convolution with three iterations by 0.6 AP, indicating a stronger deforming ability. We also find that adding more iterations does not necessarily improve the performance, which shows that it might be harder to train the network with more iterations.}
\vspace{-3mm}
\label{table:iterative}
\end{table}

To validate the advantages of our network, the model in the second row keeps the convolution operator as graph convolution and replaces Graph-ResNet with our proposed architecture, which yields 1.4 AP$_{vol}$ improvement. The main difference between the two networks is that our architecture appends a global fusion block before the prediction head.

When exploring the influence of the contour initialization, we add the initial contour proposal before the contour deformation. Instead of directly using the ellipse, the proposal step generates an octagon initialization by predicting four object extreme points, which not only compensates for the detection errors but also encloses the object more tightly. The comparison between the second and the third row shows a 1.3 improvement in terms of AP$_{vol}$.

\begin{figure}[t]
\centering
\includegraphics[width=1\linewidth]{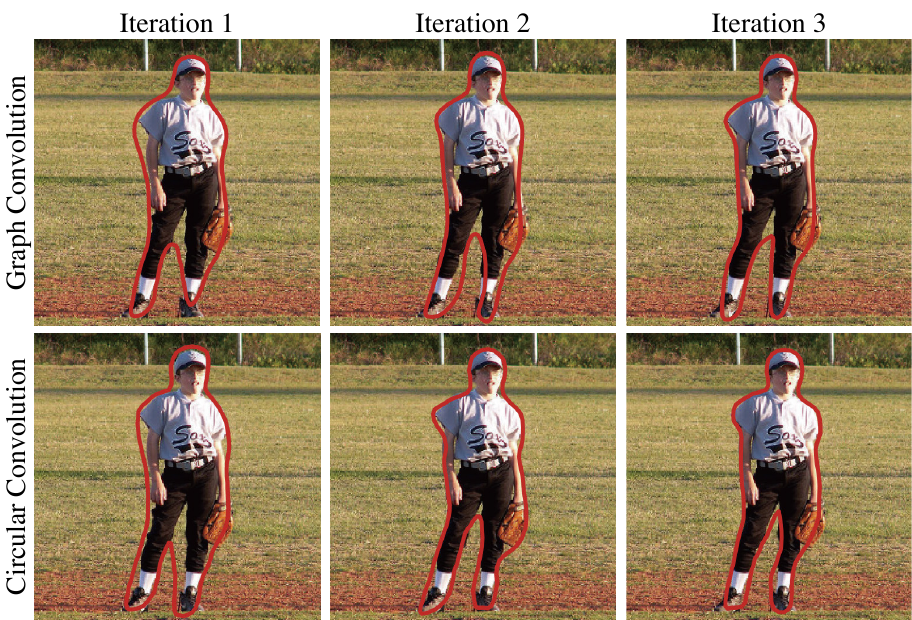}
\caption{\textbf{Comparison between graph convolution (top) and circular convolution (bottom) on SBD.} The result of circular convolution with two iterations is visually better than that of graph convolution with three iterations.}
\label{fig:iterative_inference}
\vspace{-2mm}
\end{figure}

Finally, the graph convolution is replaced with the circular convolution, which achieves 0.8 AP$_{vol}$ improvement. To fully validate the importance of circular convolution, we further compare models with different convolution operators and different inference iterations, as shown in table~\ref{table:iterative}. Circular convolution outperforms graph convolution across all inference iterations. Circular convolution with two iterations outperforms graph convolution with three iterations by 0.6 AP$_{vol}$. Figure~\ref{fig:iterative_inference} shows qualitative results of graph and circular convolution on SBD, where circular convolution gives a sharper boundary. Both the quantitative and qualitative results indicate that models with the circular convolution have a stronger ability to deform contours.

\subsection{Comparison with the state-of-the-art methods}

\begin{figure*}[t]
\centering
\includegraphics[width=1\linewidth]{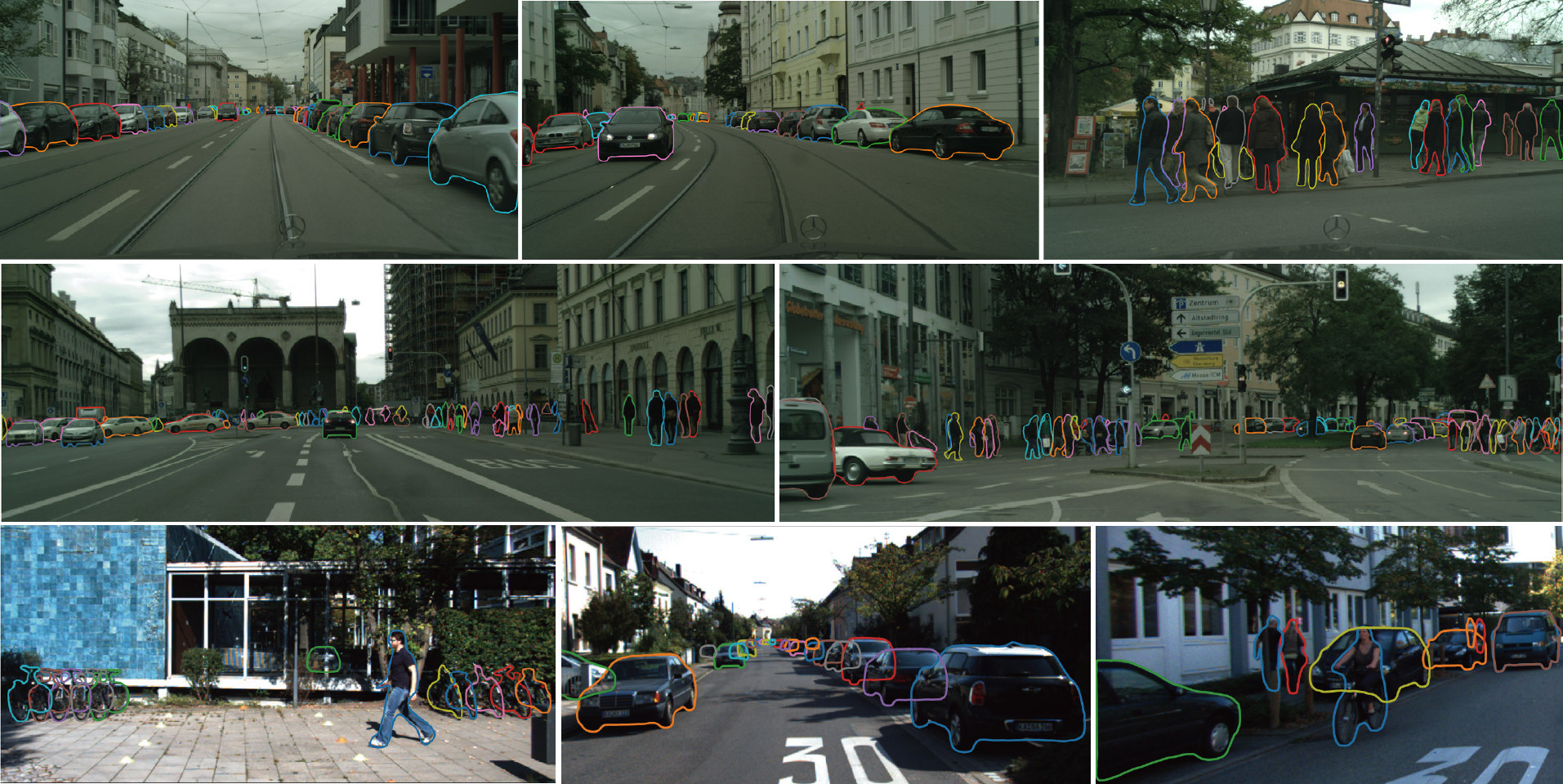}
\caption{\textbf{Qualitative results on Cityscapes test and KINS test sets.} The first two rows show the results on Cityscapes, and the last row lists the results on KINS. Note that the results on KINS are for amodal instance segmentation.}
\label{fig:qualitative_result}
\vspace{-2mm}
\end{figure*}

\begin{table*}[t]
\tablestyle{4pt}{1.05}
\begin{tabular}{l|l|x{18}|x{32}|x{18}x{18}|x{22}x{22}x{22}x{22}x{22}x{22}x{22}x{22}}
 & \multicolumn{1}{c|}{training data} & fps & AP [\texttt{val}] & AP & AP$_{50}$
 & person & rider & car & truck & bus & train & mcycle & bicycle \\[.1em]
\shline
SGN \cite{liu2017sgn} & \texttt{fine} + \texttt{coarse} & 0.6 & 29.2 & 25.0 & 44.9 & 21.8 &	20.1 &	39.4 &	24.8 &	33.2 &	30.8 &	17.7 &	12.4 \\
PolygonRNN++ \cite{acuna2018efficient} & \texttt{fine} & - & - & 25.5 & 45.5 & 29.4 & 21.8 & 48.3 & 21.1 & 32.3 & 23.7 & 13.6 & 13.6 \\
Mask R-CNN \cite{he2017mask} & \texttt{fine} & 2.2 & 31.5 & 26.2 & 49.9 & 30.5 & 23.7 & 46.9 & 22.8 & 32.2 & 18.6 & 19.1 & 16.0 \\
 GMIS \cite{liu2018affinity} & \texttt{fine} + \texttt{coarse} & - & - & 27.6 & 49.6 & 29.3 & 24.1 & 42.7 & 25.4 & 37.2 & \textbf{32.9} & 17.6 & 11.9 \\
 Spatial \cite{neven2019instance} & \texttt{fine} & 11 & - & 27.6 & 50.9 & 34.5 & 26.1 & 52.4 & 21.7 & 31.2 & 16.4 & 20.1 & 18.9 \\
 PANet \cite{liu2018path} & \texttt{fine} & $<$1 & 36.5 & \textbf{31.8} & 57.1 & 36.8 & \textbf{30.4} & 54.8 & 27.0 & 36.3 & 25.5 & \textbf{22.6} & \textbf{20.8} \\
\hline
 Deep snake & \texttt{fine} & 4.6 & \textbf{37.4} & 31.7 & \textbf{58.4} & \textbf{37.2} & 27.0 & \textbf{56.0} & \textbf{29.5} & \textbf{40.5} & 28.2 & 19.0 & 16.4 \\
\end{tabular}\vspace{1mm}
\caption{\textbf{Results on Cityscapes val (``AP [\texttt{val}]'' column) and test (remaining columns) sets.} Our approach achieves the state-of-the-art performance, which outperforms PANet \cite{liu2018path} by 0.9 AP on the val set and 1.3 AP$_{50}$ on the test set. In terms of the inference speed, our approach is approximately five times faster than PANet. The timing results of other methods were obtained from \cite{neven2019instance}.}\vspace{-3mm}
\label{table:cityscapes}
\end{table*}

\label{section:performance}

\paragraph{Performance on Cityscapes.} Since fragmented instances are very common in Cityscapes, the proposed multi-component detection strategy is adopted. Our network is trained with multi-scale data augmentation and tested at a single resolution of $1216 \times 2432$. No testing tricks are used. The detector is first trained alone for 140 epochs, and the learning rate starts from $1e^{-4}$ and drops by half at 80, 120 epochs. Then the detection and snake branches are trained end-to-end for 200 epochs, and the learning rate starts from $1e^{-4}$ and drops by half at 80, 120, 150 epochs. We choose a model that performs best on the validation set.

Table~\ref{table:cityscapes} compares our results with other state-of-the-art methods on the Cityscapes validation and test sets. All methods are tested without tricks. Using only the fine annotations, our approach achieves state-of-the-art performances on both validation and test sets. We outperform PANet by 0.9 AP on the validation set and 1.3 AP$_{50}$ on the test set. 
Our approach achieves 28.2 AP on the test set when the strategy of handling fragmented instances is not adopted. Visual results are shown in Figure~\ref{fig:qualitative_result}.

\paragraph{Performance on KINS.} The KINS dataset is for amodal instance segmentation, where objects are all annotated as single-component, so the multi-component detection strategy is not adopted. We train the detector and snake end-to-end for 150 epochs. The learning rate starts from $1e^{-4}$ and decays with 0.5 and 0.1 at 80 and 120 epochs, respectively. We perform multi-scale training and test the model at a single resolution of $768 \times 2496$.

Table~\ref{table:kins} shows the comparison with \cite{dai2016instance, li2017fully, follmann2019learning, he2017mask, liu2018path} on the KINS dataset in terms of the AP metric. 
Our approach achieves the best performance across all methods.
We find that the snake branch can improve the detection performance. When CenterNet is trained alone, it obtains 30.5 AP on detection. When trained with the snake branch, its performance improves by 2.3 AP.
For an image resolution of $768 \times 2496$ on the KINS dataset, our approach runs at 7.6 fps on a 1080 Ti GPU. Figure~\ref{fig:qualitative_result} shows some qualitative results on KINS.

\begin{figure*}[t]
\centering
\includegraphics[width=1\linewidth]{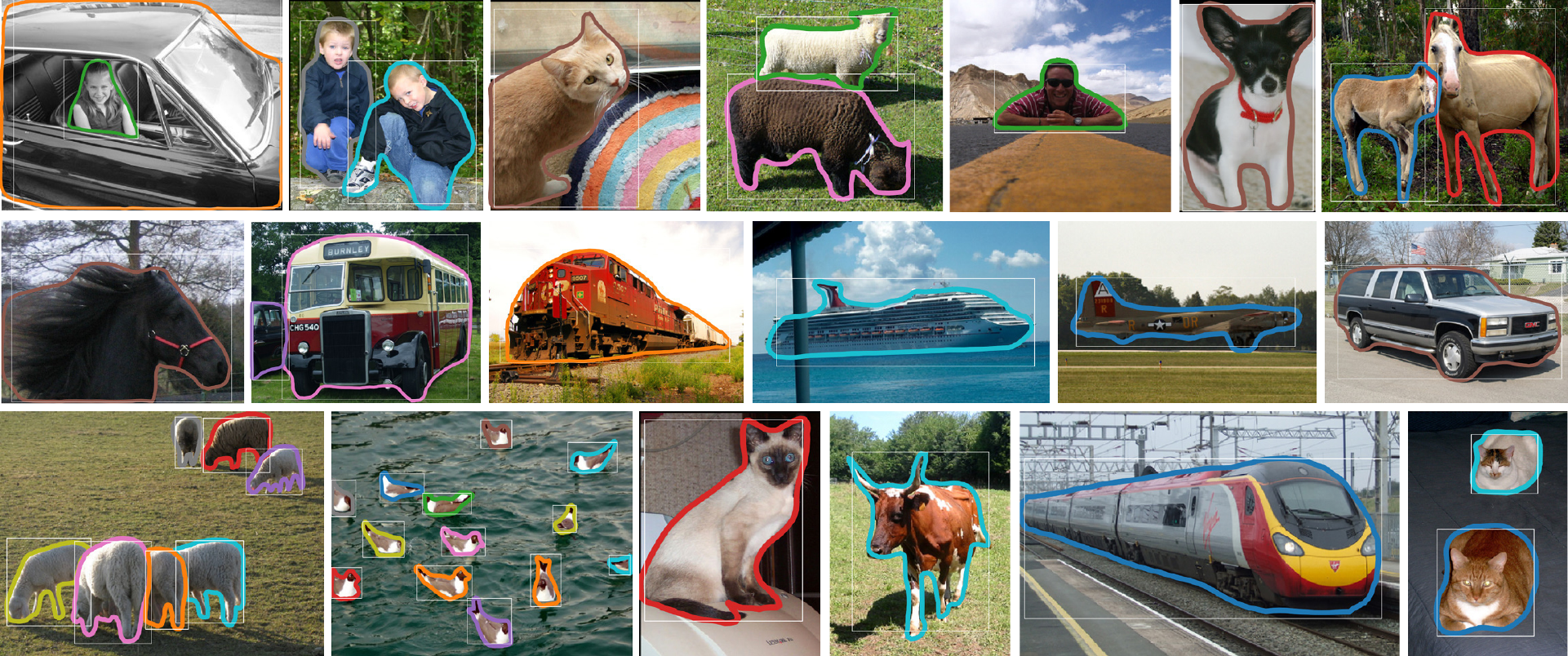}
\caption{\textbf{Qualitative results on SBD val set.} Our approach handles errors in object localization in most cases. For example, in the first image, although the detected box doesn't fully enclose the car, our approach recovers the complete car shape. Zoom in for details.}
\label{fig:sbd_qualitative_result}
\vspace{-2mm}
\end{figure*}

\begin{table}
\tablestyle{4pt}{1.05}
\begin{tabular}{l|x{42}|x{42}|x{42}}
 & detection & amodal seg & inmodal seg \\[.1em]
\shline
MNC \cite{dai2016instance} & 20.9 & 18.5 & 16.1 \\
FCIS \cite{li2017fully} & 25.6 & 23.5 & 20.8 \\
ORCNN \cite{follmann2019learning} & 30.9 & 29.0 & 26.4 \\
Mask R-CNN \cite{he2017mask} & 31.1 & 29.2 & $\times$ \\
Mask R-CNN \cite{he2017mask} & 31.3 & 29.3 & 26.6 \\
PANet \cite{liu2018path} & 32.3 & 30.4 & 27.6 \\
\hline
Deep snake & \textbf{32.8} & \textbf{31.3} & $\times$ \\
\end{tabular}\vspace{1mm}
\caption{\textbf{Results on KINS test set in terms of the AP metric.} The amodal bounding box is used as the ground truth in the detection task. $\times$ means no such output in the corresponding method.}
\vspace{-3mm}
\label{table:kins}
\end{table}

\paragraph{Performance on SBD.} Since annotations of objects on SBD are mostly single-component, the multi-component detection strategy is not adopted. For fragmented instances, our approach detects their components separately instead of detecting the whole object. We train the detection and snake branches end-to-end for 150 epochs with multi-scale data augmentation. The learning rate starts from $1e^{-4}$ and drops by half at 80 and 120 epochs. The network is tested at a single scale of $512 \times 512$.

\begin{table}
\tablestyle{4pt}{1.05}
\begin{tabular}{l|x{42}|x{42}|x{42}}
& AP$_{vol}$ & AP$_{50}$ & AP$_{70}$ \\[.1em]
\shline
STS \cite{jetley2017straight} & 29.0 & 30.0 & 6.5 \\
ESE-50 \cite{xu2019explicit} & 32.6 & 39.1 & 10.5 \\
ESE-20 \cite{xu2019explicit} & 35.3 & 40.7 & 12.1 \\
\hline
Deep snake & \textbf{54.4} & \textbf{62.1} & \textbf{48.3} \\
\end{tabular}\vspace{1mm}
\caption{\textbf{Results on SBD val set.} Our approach outperforms other contour-based methods by a large margin. The improvement increases with the IoU threshold: 21.4 in AP$_{50}$ and 36.2 in AP$_{70}$.}
\vspace{-3mm}
\label{table:sbd}
\end{table}

In Table~\ref{table:sbd}, we compare with other contour-based methods \cite{jetley2017straight, xu2019explicit} on the SBD dataset in terms of the VOC AP metrics.
\cite{jetley2017straight, xu2019explicit} predict the object contours by regressing shape vectors. 
STS \cite{jetley2017straight} defines the object contour as a radial vector, and ESE \cite{xu2019explicit} approximates object contour with the Chebyshev polynomial.
We outperform these methods by a large margin of at least 19.1 AP$_{vol}$. Note that, our approach yields 21.4 AP$_{50}$ and 36.2 AP$_{70}$ improvements, demonstrating that the improvement increases as the IoU threshold gets smaller. This indicates that our method outlines object boundaries more precisely. For $512 \times 512$ images on the SBD dataset, our approach runs at 32.3 fps on a 1080 Ti. Some qualitative results are illustrated in Figure~\ref{fig:sbd_qualitative_result}.

\paragraph{Performance on COCO.} Similar to the experiment on SBD, the multi-component detection strategy is not adopted. The network is trained with multi-scale data augmentation and tested at the original image resolution without tricks (e.g., flip augmentation). The detection and snake branches are trained end-to-end for 160 epochs, where the detector is initialized with the pretrained model released by \cite{zhou2019objects}. The learning rate starts from $1e^{-4}$ and drops by half at 80 and 120 epochs. We choose a model that performs best on the validation set. Table~\ref{table:coco_result} compares our method with other real-time methods. Our method achieves 30.3 segm AP and 33.2 bbox AP on COCO test-dev set with 27.2 fps.

\begin{table}
\begin{center}
\scalebox{0.8}{
\begin{tabular}{c|ccc}
 & YOLACT \cite{bolya2019yolact} & ESE \cite{xu2019explicit} & OURS \\[.1em]
\shline
val (segm AP) & 29.9 & 21.6 & \textbf{30.5} \\ \hline
test-dev (segm AP) & 29.8 & - & \textbf{30.3}
\end{tabular}
} 
\vspace{-2mm}
\end{center}
\caption{\textbf{Comparison with other real-time methods} on COCO.}
\vspace{-4mm}
\label{table:coco_result}
\end{table}

\subsection{Running time}

Table~\ref{table:running_time} compares our approach with other methods \cite{dai2016instance, li2017fully, he2017mask, jetley2017straight, xu2019explicit} in terms of running time on the PASCAL VOC dataset. Since the SBD dataset shares images with PASCAL VOC, the running time on the SBD dataset is technically the same as the one on PASCAL VOC. We obtain the running time of other methods from \cite{xu2019explicit}.

For $512 \times 512$ images on the SBD dataset, our algorithm runs at 32.3 fps on a desktop with an Intel i7 3.7GHz and a GTX 1080 Ti GPU, which is efficient for real-time instance segmentation. Specifically, CenterNet takes 18.4 ms, the initial contour proposal takes 3.1 ms, and each iteration of contour deformation takes 3.3 ms. Since our approach outputs the object boundary, no post-processing like upsampling is required. If the multi-component detection strategy is adopted, the detector additionally takes 3.6 ms.

\begin{table}
\tablestyle{4pt}{1.05}
\begin{tabular}{c|x{20}x{20}x{20}x{20}x{20}x{20}}
method & MNC & FCIS & MS & STS & ESE & OURS \\[.1em]
\shline
time (ms) & 360 & 160 & 180 & 27 & 26 & 31 \\
\hline
fps & 2.8 & 6.3 & 5.6 & 37.0 & 38.5 & 32.3 \\
\end{tabular}\vspace{1mm}
\caption{\textbf{Running time on the PASCAL VOC dataset.} ``MS'' represents Mask R-CNN \cite{he2017mask} and ``OURS'' represents our approach. The last three methods are contour-based methods.}
\vspace{-3mm}
\label{table:running_time}
\end{table}

\section{Conclusion}

We proposed a learning-based snake algorithm for real-time instance segmentation, which introduces the circular convolution for efficient feature learning on the contour and regresses vertex-wise offsets for the contour deformation.
Based on deep snake, we developed a two-stage pipeline for instance segmentation: initial contour proposal and contour deformation.
We showed that this pipeline gained a superior performance than direct regression of the coordinates of the object boundary points.
To overcome the limitation of the contour representation that it can only outline one connected component, we proposed the multi-component detection strategy and demonstrated the effectiveness of this strategy on Cityscapes. The proposed model achieved competitive results on the Cityscapes, Kins, Sbd and COCO datasets with a real-time performance.

\vspace{1em}
\noindent\textbf{Acknowledgements:}
The authors would like to acknowledge support from NSFC (No. 61806176) and Fundamental Research Funds for the Central Universities (2019QNA5022).

{\small
\bibliographystyle{ieee_fullname}
\bibliography{egbib}
}

\end{document}